\documentclass[sigconf,nonacm]{acmart}
\usepackage{graphicx} % Required for inserting images

\usepackage{url}            % simple URL typesetting
\usepackage{parskip}
\usepackage[utf8]{inputenc} % allow utf-8 input
\usepackage[T1]{fontenc}    % use 8-bit T1 fonts
\usepackage{xcolor}  
\usepackage{hyperref}       % hyperlinks

\usepackage{booktabs}       % professional-quality tables
\usepackage{amsfonts}       % blackboard math symbols
\usepackage{nicefrac}       % compact symbols for 1/2, etc.
\usepackage{microtype}      % microtypography
\usepackage{amsmath}
\usepackage{mathtools}
\usepackage{amsthm, bbm, bm}
\usepackage{thm-restate, thmtools, enumitem}

\usepackage{algorithm}
\usepackage{soul, enumitem}
\usepackage{multicol}
 \usepackage{algpseudocode}

\newtheorem{assumption}{Assumption}
\newtheorem{definition}{Definition}

\newtheorem{remark}{Remark}
\newtheorem{prop}{Proposition}

\usepackage[normalem]{ulem}
\usepackage[capitalize,noabbrev]{cleveref}

\usepackage{subfigure}
\DeclarePairedDelimiterX\brk[2]{\langle}{\rangle}{#1\,,\,#2} % pairing notation
\DeclarePairedDelimiterX\Set[2]{\{}{\}}{#1 \;;\; #2} % set notation

\newcommand{\argmax}{\operatornamewithlimits{arg\,max}}

\newcommand\independent{\protect\mathpalette{\protect\independenT}{\perp}}
\def\independenT#1#2{\mathrel{\rlap{$#1#2$}\mkern2mu{#1#2}}}
\newcommand\indep\independent

\newcommand{\EE}{\mathbb{E}}

\newcommand{\PP}{\mathbb{P}}

\newcommand{\RR}{\mathbb{R}}

\newcommand{\Acal}{\mathcal{A}}

\newcommand{\Ccal}{\mathcal{C}}

\newcommand{\Pcal}{\mathcal{P}}

\newcommand{\Xcal}{\mathcal{X}}
\newcommand{\Ycal}{\mathcal{Y}}

\newcommand{\Bal}{\begin{align}}
\newcommand{\Eal}{\end{align}}
\newcommand{\Beq}{\begin{equation}}
\newcommand{\Eeq}{\end{equation}}
\newcommand{\Bit}{\begin{itemize}}
\newcommand{\Eit}{\end{itemize}}
\newcommand{\Ben}{\begin{enumerate}}
\newcommand{\Een}{\end{enumerate}}
\newcommand{\Ba}{\begin{array}}
\newcommand{\Ea}{\end{array}}

\newcommand{\Bvec}{\left(\begin{array}{c}}
\newcommand{\Evec}{\end{array}\right)}

\newcommand{\Bmat}{\left(\begin{array}}
\newcommand{\Emat}{\end{array}\right)}

\newcommand{\Bol}{\begin{outline}}
\newcommand{\Eol}{\end{outline}}

\title[CSPI-MT: Calibrated Safe Policy Improvement with Multiple Testing for Threshold Policies]{CSPI-MT: Calibrated Safe Policy Improvement \\ with Multiple Testing for Threshold Policies}
\author{Brian M Cho*}
\affiliation{
\institution{Cornell University}
\titlenote{Work done during an internship at Meta.}
\country{New York, NY}
}

\author{Ana-Roxana Pop}
\affiliation{
\institution{Meta}
\country{New York, NY}
}

\author{Kyra Gan}
\affiliation{
\institution{Cornell University}
\country{New York, NY}
}

\author{Sam Corbett-Davies}
\affiliation{
\institution{Meta}
\country{Menlo Park, CA}
}

\author{Israel Nir}
\affiliation{
\institution{Meta}
\country{San Francisco, CA}
}
\author{Ariel Evnine}
\affiliation{
\institution{Meta}
\country{San Francisco, CA}
}

\author{Nathan Kallus}
\affiliation{
\institution{Cornell University, Netflix}
\country{New York, NY}
}

\date{August 8, 2024}

\begin{document}

\begin{abstract}
    When modifying existing policies in high-risk settings, it is often necessary to ensure with high certainty that the newly proposed 
    policy improves upon a baseline, such as the status quo. 
    In this work, we consider the problem of \emph{safe policy improvement}, where one only adopts a new policy if it is deemed to be better than the specified baseline with at least pre-specified probability. We focus on threshold policies, a ubiquitous class of policies with applications in economics, healthcare, and digital advertising. 
    Existing methods rely on potentially underpowered safety checks and limit the opportunities
    for finding safe improvements, so too often they must revert to the baseline to maintain safety. 
    We overcome these issues by leveraging the most powerful safety test in the asymptotic regime and allowing for  multiple candidates 
    to be tested
    for improvement over the baseline. 
    We show that in adversarial settings, our approach controls the rate of adopting a policy 
    worse than 
    the baseline to the pre-specified
    error level, even in moderate sample sizes. 
    We present 
    CSPI and CSPI-MT, 
    two novel heuristics for 
    selecting
    cutoff(s) to maximize the policy improvement from baseline.
    We demonstrate through both synthetic and external datasets that our approaches improve both the detection rates of safe policies and the realized improvement, 
    particularly under stringent safety requirements and low signal-to-noise conditions. 
\end{abstract}

\maketitle

\section{Introduction}

Many policy improvement algorithms 
% are often 
focus on maximizing the expected gain in policy, 
often without providing strong
% with minimal 
probabilistic
guarantees that the policy will improve upon the baseline or status quo with high confidence \cite{athey2020policylearningobservationaldata, kitagawa, wang2017optimaladaptiveoffpolicyevaluation}. 
In high-risk settings, such as clinical trials or policy-making, these guarantees are crucial.
For digital marketers, implementing
bad policies 
can be costly, leading to unnecessary changes to existing policies accompanied by losses in performance, worse product experiences or detrimental outcomes.
To address this, multiple approaches in the literature have focused on the problem of \emph{safe policy improvement}, where a policy is adopted only if it is deemed better than the baseline, with the probability of error controlled at a specified rate \cite{laroche2019safepolicyimprovementbaseline, pmlr-v37-thomas15, scholl2022safepolicyimprovementapproaches, NIPS2003_300891a6}. 
The chosen error rate reflects the acceptable level of risk for the specific application.

For treatments with heterogeneous effects, decision-makers often leverage data collected from either randomized experiments (e.g., A/B tests) or observational studies to identify personalized treatment policies. A fundamental goal when deciding to roll out a policy change is to ensure the new proposed policy improves upon the status quo with high confidence. In this work, we focus on threshold policies, a widely-used class of policies where the targeted population is decided based on a cutoff applied to a scoring function. Common examples of score variables include estimated treatment effects \cite{threshold_ite, fernándezloría2022comparisonmethodstreatmentassignment} in digital marketing
or
risk level estimates \cite{benmichael2022safe, liver_threshold}. In industry applications, threshold policies have been used for ad assignment,  based on click-through rates \cite{ouyang2019clickthroughratepredictionuser} or estimated causal effects \cite{fernándezloría2022comparisonmethodstreatmentassignment}. Outside of industry settings, applications include using threshold policies for healthcare referrals \cite{liver_threshold} and legal recommendations \cite{benmichael2022safe}. While simply enacting policy changes based on estimated gains can lead to high chances of moving to a policy worse than the status quo, existing methods 
in the literature 
often rely on
% suffer from 
overly stringent safety checks and  
limited heuristics for testing potentially safe candidates \cite{scholl2022safepolicyimprovementapproaches}.  This results in low success rates in identifying policies that outperform the baseline and limited policy improvement.

We propose CSPI and CSPI-MT, two novel heuristics for selecting a single cutoff and multiple cutoffs, respectively, in a computationally tractable and statistically valid manner. Both algorithms employ asymptotically calibrated tests and improve both 1) the detection rates of policies that are truly better than the baseline and 2) the overall gain in policy value. We empirically test our approach using synthetic data and an external dataset involving French job-search counseling services \cite{kallus2022treatmentrisk, behaghel}. 
Our empirical results 
show that
our algorithms calibrate the empirical error rate to the desired error guarantee, detect policies that improve upon the baseline at a higher rate, and 
achieve better policy gains compared to existing approaches, especially under stringent safety requirements and noisy, low-signal settings. 

The paper is organized as follows: Section \ref{sec:setup} introduces our setting, assumptions, and problem formulation. Section \ref{sec:related_work} reviews related works,with a focus on the High Confidence Policy Improvement (HCPI) algorithm \cite{pmlr-v37-thomas15}. Utilizing sampling splitting, we first provide an asymptotically optimal testing procedure of a single cutoff on the testing dataset in Section~\ref{sec:testing_procedures}, and then extend it
to handle multiple cutoff policies. In Section \ref{sec:cutoff_selection}, we provide heuristic algorithms for selecting single and multiple cutoffs using a small tuning dataset. We present our heuristics in Algorithm~\ref{alg:MDR-HCPI} in Section~\ref{subsec:adaptive_multi_cutoffs}. Finally, we numerically test our procedures in Section \ref{sec:empirics}.

\section{Statistical Setup and Notation}\label{sec:setup}

\subsection{Assumptions and Notation}
We observe $n$ independent and identically distributed tuples $\{O_i\}_{i = 1}^n = \{(X_i, A_i, Y_i)\}_{i=1}^n$, where $X \in \Xcal \subset \RR^d$ denotes pretreatment covariates/features with dimension $d$, $A \in \{0,1\}$ is the treatment indicator (with $A=1$ indicating that the unit received treatment), and $Y \in \Ycal \subset \RR$ is the primary 
% metric/
outcome of interest.\footnote{We do not impose additional restrictions on $\Xcal$ and $\Ycal$, which may consist of discrete, continuous, or mixed values.} We assume the following conditions on the data generating process (DGP) $P$.

\begin{assumption}[Potential Outcomes Model \cite{rubin_po_model}]\label{assump:rubin_po}
    % We assume that
    For each unit $i$,
    % (X_i, A_i, Y_i)$, 
    there exist two potential outcomes $(Y_i(0), Y_i(1))$ and $Y_i = Y_i(A_i)$. 
\end{assumption}

Assumption \ref{assump:rubin_po} implies that the observed outcomes of each unit only depend on that unit's treatment, 
with no interference between units.
We make the standard unconfoundedness and positivity assumptions that enable the estimation of the policy values/differences:

\begin{assumption}[Unconfoundedness] \label{assump:unconfoundedness}
    % We assume that 
    The treatment assignment in our observed data only depends on $X$, i.e., $\{Y_i(0), Y_i(1)\} \perp A_i | X_i$. 
\end{assumption}
Let $e(X) \coloneqq P(A = 1|X=x)$ be the propensity score, denoting the probability of treatment given pretreatment covariates $X$.
\begin{assumption}[Positivity] \label{assump: positivty}
For all $x \in \Xcal$,
     % We assume that
     $c \leq e(X) \leq 1-c $ for some constant $0 < c < \frac{1}{2}$. 
     % for all $x \in \Xcal$. 
\end{assumption}

Assumption \ref{assump:unconfoundedness} states that the treatment assignment mechanism for the observed data only depends on the observable characteristics: this is trivially satisfied in 
randomized experiments
where the treatment assignment policy is a known function of $X$. 
Assumption \ref{assump:unconfoundedness} in general does not hold in observational studies and must be carefully considered. 
The positivity assumption is necessary for identification -- without it, estimation and inference with our desired statistical tests is impossible, even with an infinite amount of data. 
We denote the set of all DGPs $P$ that satisfy Assumptions \ref{assump:rubin_po}, \ref{assump:unconfoundedness}, \ref{assump: positivty} as the set of distributions $\Pcal$. 

Let $\pi: \Xcal \rightarrow \{0,1\}$ denote a given treatment assignment policy.

\begin{definition}[Policy Value]
The value of a policy, $V(\pi)$, is the average outcome of interest under 
$\pi$:   $V(\pi) \coloneqq \EE[Y(\pi(X))]$. 
\end{definition}

Lastly, we define the threshold policy class, a set of policies widely used in high-risk settings where safety guarantees are desired.

\begin{definition}[Threshold Policy Class]
    Let $g:\Xcal \rightarrow \RR$ be a known scoring function, and let $S  = g(X)$. We define the class of threshold policies, $\Pi$, with respect to $S$ as
    $\Pi
    (S) = \{ \mathbf{1}[S \geq c] \}_{c \in \RR}.$
    We define an instance from this policy class, $\mathbf{1}[S \geq c] $, as $\pi(c)$. 
\end{definition}

We denote the cutoff of the baseline policy as $c_0$, 
and the
corresponding policy 
as 
$\pi(c_0)$. Natural choices of $c_0$ include $c_0 = \infty$, which corresponds to treating no one in the population, and $c_0 = -\infty$, which corresponds to treating the entire population.

\subsection{Problem Statement}

Given i.i.d. observations $\{O_i\}_{i=1}^n$,  error tolerance bound $\gamma$, 
a discrete candidate set of cutoffs $\Ccal = \{c_1, ..., c_G\} \in \RR^G$, and a pre-specified baseline policy $\pi(c_0)$, we say a policy improvement algorithm, 
$\Acal:\left(\{O_i\}_{i=1}^n, \gamma, \Ccal, 
\pi(c_0)\right) \rightarrow \hat{c} \in \Ccal$, is
\emph{asymptotically safe}\footnote{We defer the discussion of
our rationale for asymptotic guarantees to the next section.} if it satisfies the following:

\begin{definition}[Asymptotic Safety]\label{defn:asymp_safety}
    As $n$ grows to infinity, the probability that the selected policy  $\pi(\hat{c})$
    will have a value lower than the baseline policy, $V(\pi(\hat{c})) < V(\pi(c_0))$,
    is at most $\gamma$, 
    % as $n$ grows large,
    i.e., 
    $$ \lim_{n \rightarrow \infty} \sup_{P \in \Pcal} \PP_P\left(V(\pi(\hat{c})) < V(\pi(c_0) \right) \leq \gamma,$$
    where $\PP_P(\cdot)$ denotes the probability measure with respect to $P$.
\end{definition}
Definition~\ref{defn:asymp_safety} requires that
the safety guarantee holds 
across all distributions $\Pcal$ that satisfy our assumptions, regardless of how the policy $\pi(\hat{c})$ is selected. 
We additionally 
note that the policy values themselves are not random. The uncertainty in our probabilistic statement comes from the randomness in the observed data $\{O_i\}_{i=1}^n$ and the method used to select a proposed cutoff based on this data. 

Given an asymptotically safe algorithm, we aim to (1) optimize calibration across any distribution $P \in \Pcal$, and (2) maximize the realized policy improvement.
We define \emph{calibration error}
with respect to the pre-specified error rate $\gamma$ 
as below:
\begin{definition}[Calibration Error]
    The calibration error of an algorithm $\Acal$, denoted $\beta_n(\Acal)$, is 
    the difference between the pre-specified error rate $\gamma$ and the error rate across distributions $P \in \Pcal$.  
    $$\beta_n(\Acal, P) = \left|\PP_P\left(V(\pi(\hat{c})) < V(\pi(c_0))\right) - \gamma\right|.$$
\end{definition}
We note that calibration error is inherently a finite-sample quantity that measures the deviation of the true error rate of algorithm $\Acal$ (i.e., the probability of selecting a policy with a worse value) 
from the desired error rate $\gamma$ 
when using $n$ samples. 

Lastly, the primary objective of any policy improvement algorithm is to improve the policy value from the baseline. We define the \emph{expected improvement} of an algorithm $\Acal$ from the baseline as follows:

\begin{definition}[Expected Improvement]
    Given a distribution $P \in \Pcal$, and a policy improvement algorithm $\Acal$,
    we define the expected improvement of algorithm $\Acal$ as follows:
    $$\tau\left(\Acal, n, \gamma, \Ccal, \pi(c_0)\right) = \EE_P\left[V(\pi(\hat{c})) - V(\pi(c_0))\right].$$
\end{definition}
In this paper, we introduce CSPI and CSPI-MT, two asymptotically safe policy improvement algorithms 
designed
to improve detection power and 
increase expected improvement 
compared to
existing approaches in the literature. 
We leverage \emph{the most 
powerful
asymptotic
tests} based on limiting normal distributions, and allowing our method to 
test multiple policy candidates through \emph{simultaneous confidence bands}. 
Our approaches 1) require minimal hyperparameter tuning beyond the necessary inputs, 2) work well with a large number of candidate policies, and 3) significantly improve the practical performance compared with existing approaches. 

Below, we briefly 
discuss
additional practical applications that directly 
align with
our objectives.

\paragraph{Application 1: Identifying Negatively Affected 
Cohorts.} An equivalent interpretation of our problem, when combined with threshold policies, is 
identifying
the set of 
individuals
with scores $S$ below a given threshold $c$ 
who are statistically significantly affected, either negatively or positively.
For example, with the baseline of treat all, detecting a policy $\pi(c) = \{\mathbf{1}[S \geq c]\}$ indicates that the 
subgroup 
with $S < c$ 
are negatively affected, with confidence $1-\gamma$.
Conversely,
with the baseline of treat none, detecting the same policy
indicates that the subgroup 
with $S \geq c$ are positively affected with confidence $1-\gamma$.

\paragraph{Application 2: Assessing Pre-Trained Scores for Downstream Treatment Delivery.} In many cases \cite{benmichael2022safe, liver_threshold}, the score variable $S$ is obtained through a carefully designed statistic that roughly corresponds to how well an individual is likely to benefit from the treatment option, with lower scores representing worse outcomes under treatment relative to control. This statistic $S$ can be designed through domain expertise or pre-trained machine-learned treatment effect predictions on a holdout dataset. 
In settings where $S$ denotes individual-level treatment effect (ITE) estimates, the natural baseline policy is to treat positively affected individuals, i.e., with $c_0 = 0$. Our procedure can directly be used to both assess ITE model calibration 
and identify the threshold cutoff that will yield improved policy value based on the current model. For example, finding a cutoff $\hat{c} \neq 0$ that improves upon the baseline policy value $V(\pi(0))$ with small $\gamma$ values provides evidence that the current model may be miscalibrated. On the other hand, if the score model captures the relative ordering of the ITE but provides imprecise ITE values, our approach offers a principled way to recalibrate the policy cutoff using these pre-trained scores $S$ based on statistical significance.

Next, we review related work, providing 1) the framework for ensuring statistical validity, 2) the details on our choice of asymptotic safety guarantees, and 3) related background on asymptotically powerful tests.

\section{Related Work} \label{sec:related_work}

Our work builds upon three 
streams of literature: (1) data splitting to avoid hypothesis fishing
or $p$-hacking, (2) safe policy improvement in reinforcement learning (RL), 
and (3) asymptotically optimal estimators and tests from the causal inference literature.

\paragraph{The problem of hypothesis fishing.} A naïve approach to picking a safe policy would be to construct standard Wald lower confidence bounds on the  differences between $\{V(\pi(c))\}_{c \in \Ccal}$ and $V(\pi (c_0))$ using the entire data $\{O_i\}_{i=1}^n$, and selecting a policy $\pi(\hat{c})$ whose lower bound is above 0. 
However, without proper correction via simultaneous bands \cite{sup_t_band} or uniform concentration \cite{cho2024peekingpeaksequentialnonparametric}, this approach will inflate error rates far beyond the specified $\gamma$-level \cite{multi_test_bad}, a problem also known as \emph{hypothesis fishing}. A general approach to avoid this problem 
is to randomly split the observed sample $\{O_i\}_{i=1}^n$ into two disjoint sets $D_{\text{tune}}$ and $D_{\text{test}}$ \cite{cox1975note, rubin2006method, ignatiadis2016data, pmlr-v37-thomas15}, where $D_{\text{tune}}$ is used 
to determine 
candidate policies,
and
$D_{\text{test}}$ is used to test
the selected 
candidates. To avoid significant loss of power,
$D_{\text{tune}}$ is often chosen to be far smaller
than $D_{\text{test}}$.
For example, in HCPI \cite{pmlr-v37-thomas15}, $D_{\text{tune}}$ only has 20\% of the total number of samples.
Our approach falls under this general framework and improves both the selection of tested cutoffs in $D_{\text{tune}}$ 
and the detection power in $D_{\text{test}}$, 
tailored for
safe policy improvement 
in threshold policies.

\paragraph{Safe policy improvement approaches.}
Safe policy improvement approaches have been studied in the 
RL 
setting \cite{scholl2022safepolicyimprovementapproaches, laroche2019safepolicyimprovementbaseline, pmlr-v37-thomas15}. The closest paper to ours is the High-Confidence Policy Improvement (HCPI) algorithm \cite{pmlr-v37-thomas15}, which addresses the same problem but focuses only on testing a \emph{single} cutoff. Despite the similar setup, HCPI differs significantly from our algorithm. First, HCPI guarantees improved performance over a baseline policy value, which can introduce additional uncertainty when estimating this value in practice \cite{laroche2019safepolicyimprovementbaseline}. HCPI does not account for this uncertainty and cannot be directly applied to settings where the baseline policy value is uncertain.
In contrast, our approach works directly with the baseline policy itself. Second, rather than providing multiple testing schemes that provide different safety guarantees under different conditions as in \cite{pmlr-v37-thomas15}, 
we focus on providing asymptotically efficient safety guarantees.For example, \cite{pmlr-v37-thomas15} provides finite-sample guarantees by leveraging
concentration inequalities, which
are known to be conservative \cite{austern2024efficient, scholl2022safepolicyimprovementapproaches, pmlr-v37-thomas15},
leading to 
reduced
detection power and 
policy improvement. 
Meanwhile, bootstrap-style inference methods offer well-calibrated error guarantees but are computationally inefficient. This inefficiency becomes particularly evident when the number of policies is large or when the cutoff variable is continuous, requiring discretization for cutoff search.
We further expand on our choice of asymptotic guarantees below.

\paragraph{Asymptotically efficient testing procedures.} 
To balance well-powered testing procedures with computational efficiency, we focus on characterizing the \emph{most powerful} asymptotic test (i.e., the one with the smallest asymptotic sampling variance) for a single cutoff to achieve our safety guarantees. 
This approach adapts the well-studied double machine learning (DML) approach in the causal inference literature \cite{chernozhukov2017doubledebiased, JMLR:v21:19-827} to our problem,
enabling the use of powerful machine learning methods under minimal assumptions to improve the power of our asymptotic tests. We contrast our 
approaches with the inverse-propensity-weighted (IPW) estimates used by \cite{pmlr-v37-thomas15}, and show that our methods offer better-calibrated error guarantee, detection power, and expected improvement. 

We further extend our asymptotic test for a single cutoff to \emph{multiple} cutoffs 
by leveraging the asymptotic distributions of the estimators that underlie our test.
In the setting where the variance is high and the signal is low, selecting a single cutoff
on $D_{\text{tune}}$, as in HCPI \cite{pmlr-v37-thomas15},
often leads to safety test failures, resulting in no policy changes.
We overcome this issue by using \emph{simultaneous} lower confidence bounds, enabling 
multiple cutoff testing. 
We 
show that our
heuristic algorithms 
empirically perform well across both synthetic and semi-synthetic examples.

% \section{Safe Policy Improvement for Threshold Policies} 
\section{Safety Tests for Threshold Policy Improvement}
\label{sec:testing_procedures}

We follow the sampling-splitting framework by randomly dividing 
the data $\{O_i\}_{i=1}^n$
into two sets: $D_{\text{tune}}$ and $D_{\text{test}}$, with $D_{\text{tune}}$ being much smaller than $D_{\text{test}}$.
$D_{\text{tune}}$ is used to select cutoff policies to test, while $D_{\text{test}}$ is used to test the chosen policy candidate(s) from $D_{\text{tune}}$. 
In this section, we introduce our testing procedure on $D_{\text{tune}}$. 
In Section~\ref{defn:policy_diff_est}, we first 
introduce the procedure for estimating threshold policy differences given a candidate set of cutoff values.
Next, in Section~\ref{subsec:single_cutoff} we 
provide an asymptotically optimal test when testing a single chosen cutoff policy on $D_{\text{test}}$. We further generalize our approach 
to test \emph{multiple} chosen cutoff policies on $D_{\text{test}}$ by leveraging the covariance structure of the selected cutoffs in Section~\ref{subsec:multi_cutoff}.

\subsection{Estimation of Threshold Policy Differences}\label{defn:policy_diff_est}
We provide the estimation procedure of threshold policy differences given a baseline policy $\pi(c_0)$ and a set of candidate cutoff policies 
$\{\pi(c)\}_{c \in \mathbf{c}}$, where $\mathbf{c} = \{c_1,..., c_l\}$ is a given set
of cutoffs of cardinality $l$.
Let $D$ be the input dataset of the estimation procedure (e.g., $D_{\text{train}}$ or $D_{\text{test}}$).
Let $\hat{\mu}(A, X)$ denote an estimator of the conditional outcome function $\EE[Y | A, X]$, and let $\hat{e}(X)$ denote an estimator of the propensity score.
We note that 
$\hat{\mu}, \hat{e}$ can be estimated using
standard regression methods (e.g., linear/logistic regression) and
machine learning techniques such as neural networks, random forests, or gradient-boosted trees.

Our goal is to estimate the policy difference vector $\tau(\mathbf{c}):= \tau(\mathbf{c}, c_0) = [\tau(c_1, c_0), ... , \tau(c_l, c_0)]$, where $\tau(c):=\tau(c, c_0) = V(\pi(c)) - V(\pi(c_0))$. To do so,
we first split $D$ randomly into $K$-folds: $D_1,...,D_K$. 
We  denote $\hat{\mu}_k$, $\hat{e}_k$ as the estimated conditional means and propensity scores trained on all folds except the $k$-th fold. 
For each $c \in \mathbf{c}$
and data point $(X_i, A_i, Y_i)$, we first calculate the contribution of each data point to the policy difference value by defining the following functions
$\hat{\psi}^c_1,
% (X_i, A_i, Y_i, \hat{\mu}_k, \hat{e}_k),
$ $ \hat{\psi}^c_0,
% (X_i, A_i, Y_i, \hat{\mu}_k, \hat{e}_k)
$ and $\hat{\psi}^c
% (X_i, A_i, Y_i, \hat{\mu}_k, \hat{e}_k)
$ for 
each $D_k$:
% as follows:
    \begin{align}
        &\hat{\psi}_1^c(X_i, A_i, Y_i, \hat{\mu}_k, \hat{e}_k) = Y_i \frac{A_i}{\hat{e}_k(X_i)} - \hat\mu_k(1,X_i) \frac{A_i - \hat{e}_k(X_i)}{\hat{e}_k(X_i)},\\
        &\hat{\psi}_0^c(X_i, A_i, Y_i, \hat{\mu}_k, \hat{e}_k) = Y_i\frac{1-A_i}{1-\hat{e}_k(X_i)} - \hat\mu_k(0, X_i) \frac{\hat{e}_k(X_i) - A_i}{1-\hat{e}_k(X_i)},\\
        &\hat\psi^c(X_i, A_i, Y_i, \hat{\mu}_k, \hat{e}_k) = \left(1-2(\mathbf{1}[c_0 \geq c])\right) \times \label{eq:c0sign}\\  &\quad\qquad\qquad\qquad\qquad \mathbf{1}[S_i \in (\min(c_0, c), \max(c_0, c))] \times \label{eq:SinRange} \\
        &\qquad\qquad\qquad
        \left(\hat{\psi}^c_0(X_i, A_i, Y_i, \hat{\mu}_k, \hat{e}_k) - \hat{\psi}^c_1(X_i, A_i, Y_i, \hat{\mu}_k, \hat{e}_k)\right).\label{eq:influence_fnc}
    \end{align}
    Intuitively,
    Equation~\eqref{eq:SinRange} checks if the score of the current data point, $S_i$, falls within the range where the proposed policy $\pi(c)$ differs from the baseline policy $\pi(c_0)$. This indicates whether the policy difference estimate should include the current observation $O_i = (X_i, A_i, Y_i)$; Equation~\eqref{eq:influence_fnc} calculates the contribution of each data point to the policy value by evaluating the influence function of the policy difference evaluated at the current data point. 
    Equation~\eqref{eq:c0sign} 
    aligns the sign of the increment in policy value with that of Equation~\eqref{eq:influence_fnc}, indicating that 
    the incremental policy value should be positive if the policy gain on this data point were to be positive.
    Finally, with slight abuse of notation, we define each entry of our estimated policy difference $\hat\tau(\mathbf{c}):= \hat\tau(\mathbf{c},c_0)$ and the $l\times l$ sampling covariance matrix $\widehat{\Sigma}(\mathbf{c}): = \widehat{\Sigma}(\mathbf{c}, c_0)$ as:
    % is given by the following:
    \begin{align}
        \hat\tau(c)
        % :=\hat\tau(c, c_0)
        &= \frac{1}{n}\sum_{i=1}^n \hat{\psi}^c(X_i, A_i, Y_i, \hat{\mu}_k, \hat{e}_k),\\
        \widehat{\Sigma}_{c_p,c_q}(\mathbf{c}) 
        % := \widehat{\Sigma}_{c_p,c_q}(\mathbf{c},c_0)
        &= \frac{1}{n}\sum_{i=1}^n \big(\hat\psi^{c_p}(X_i, A_i, Y_i, \hat{\mu}_k, \hat{e}_k) -\hat\tau(c_p)) \notag\\
        &\quad\quad \times (\hat\psi^{c_q}(X_i, A_i, Y_i, \hat{\mu}_k, \hat{e}_k) - \hat\tau(c_q) \big),
    \end{align}
    where $\widehat\Sigma_{c_p, c_q}(\mathbf{c})$ corresponds to the indices of cutoffs $c_p, c_q \in \mathbf{c}$ of the covariance matrix.

Under standard regularity conditions \cite{chernozhukov2017doubledebiased} provided in Proposition \ref{thm:asymp_normal_efficient}, $n^{-1/2}\left(\hat\tau(\Ccal) - \tau(\Ccal)\right)$ converges to a multivariate normal distribution $N(0, \Sigma)$, where 
$\Sigma_{c_p, c_q} = \EE_P[\hat{\psi}^{c_p}(X,A,Y, \mu, e)\hat{\psi}^{c_q}(X,A,Y, \mu, e)]$
corresponds to entries of the population covariance matrix with the true conditional outcome functions $\mu$ and propensity scores $e$. This limiting distribution underlies the asymptotic safety guarantees of our tests in Algorithms \ref{alg:safety_test} and \ref{alg:multiple_safety_test}.

\subsection{Efficient Single Cutoff Testing}\label{subsec:single_cutoff}
Once a cutoff is selected on $D_{\text{train}}$ using the above estimate, 
we provide the safety test to be performed on $D_{\text{test}}$ in Algorithm~\ref{alg:safety_test}, where
$\Phi^{-1}(\cdot)$ is the quantile function of a standard normal random variable. The following proposition provides 1) the correctness of
our asymptotic test defined in Algorithm \ref{alg:safety_test}, and
2) the conditions under which our test is optimal. Let     $\|\cdot\|_{P,2}$ denotes the $L_2$ norm with respect to distribution $P$.

\begin{algorithm}[t]
    \begin{algorithmic}[1]
    \State \textbf{Input}: $D_{\text{test}}$, baseline cutoff $c_0$, number of folds $K$, 
    % (number of folds for estimation),
    model classes for estimating $\hat{\mu}$ and $\hat{e}$, a single cutoff $\mathbf{c} = c$,
    % \in \RR$,
    error tolerance $\gamma$.

    \State Estimate $\hat\tau(c)$ and $\widehat{\Sigma}(c)$ based on Section \ref{defn:policy_diff_est} using $K$ folds on $D_{\text{test}}$.
    \State Calculate the estimated lower confidence bound:
    $$\widehat{L}(c, c_0) = \hat\tau(c) - \Phi^{-1}(1-\gamma)\sqrt{\widehat{\Sigma}(c)/|{D_{\text{test}}|}}.  $$

    \State \textbf{return} $\mathbf{1}[\widehat{L}(c,c_0) > 0]$. 
    \end{algorithmic}
    
    \caption{Safety Test on $D_{\text{test}}$ for a Single Cutoff.}
    \label{alg:safety_test}
\end{algorithm}

\begin{prop}[Asymptotic Correctness and Efficiency of Our Safety Test]\label{thm:asymp_normal_efficient}
Assume that (i) estimated functions $\hat{\mu}_k$, $\hat{e}_k$ are bounded 
with respect to $P$ almost surely, 
(ii) $\| \hat\mu_k - \mu\|_{P,2}\times \|\hat{e}_k - e \|_{P, 2} = o_P(1/\sqrt{n})$, and (iii) $\PP_P(S \in [\min(c,c_0), \max(c,c_0)])$ is bounded above 0. 
Then, our confidence intervals are asymptotically tight,
    $$\lim_{n \rightarrow \infty}\sup_{P \in \Pcal} \left|\PP_P\left( \tau(c) < \hat{\tau}(c) - \Phi^{-1}(1-\gamma) \sqrt{\widehat{\Sigma}(c)/n} \right) - \gamma   \right| = 0. $$
    % where $\|\cdot\|_{P,2}$ denotes the $L_2$ norm with respect to distribution $P$.
    % , and $\PP_P(\cdot)$ denotes the probability measure with respect to $P$.
    Furthermore, our variance estimate $\widehat{\Sigma}(c) \rightarrow {\Sigma}(c)$ converges to the minimum variance estimate among all regular and asymptotically linear estimators (see formal definition in \cite{ral_estimators}) under the set of distributions $\Pcal$ that satisfy Assumptions \ref{assump:rubin_po}, \ref{assump:unconfoundedness}, and \ref{assump: positivty}. 
\end{prop}

Proposition \ref{thm:asymp_normal_efficient} ensures that we satisfy our desired asymptotic safety guarantee in Definition \ref{defn:asymp_safety}. 
Across any choice of cutoff variable $S$ and cutoff value $c$, for any data-generating distribution $P \in \Pcal$, the safety test in Algorithm \ref{alg:safety_test} is guaranteed to provide exact $1-\alpha$ coverage as $n$ grows large. By attaining the smallest asymptotic variance among all estimators that permit normal approximation, our lower bound test in Algorithm \ref{alg:safety_test} will be the \emph{least} conservative relative to other asymptotic normal safety test approaches.

\subsection{Multiple Cutoff Testing for Robustness}\label{subsec:multi_cutoff}
While testing \emph{only} a single cutoff provides the \emph{maximal} asymptotic power for the chosen cutoff
$\hat{c}$, 
given the relative small sample size in $D_{\text{tune}}$, it may be desirable to test multiple cutoffs on $D_{\text{test}}$.
Since the selection of a single cutoff is based on a small fraction of the data, it tends to be highly unstable, regardless of how the cutoff is chosen. To test \emph{multiple cutoffs} selected from $D_{\text{tune}}$, we leverage 
the plug-in joint confidence band approach of Olea et. al. \cite{sup_t_band} and provide the pseudocode in Algorithm \ref{alg:multiple_safety_test}.

\begin{algorithm}[t]
    \begin{algorithmic}[1]
    \State \textbf{Input}: $D_{\text{test}}$, number of folds $K$,
    model classes for estimating $\hat{\mu}, \hat{e}$, a selected set of cutoffs $\mathbf{c}$,
    error tolerance $\gamma$, $n_\text{sim}$.

    \State Estimate vector and matrix $\hat{\tau}(\mathbf{c})$ and $\widehat{\Sigma}_{i,j}(\mathbf{c})$ based on Section  \ref{defn:policy_diff_est} using $K$ folds on $D_{\text{test}}$.

    \State Simulate i.i.d. normal vectors $\widehat{V}^\ell \sim N(\mathbf{0}, \widehat{\Sigma}(\mathbf{c}))$ for $\ell \in [n_{\text{sim}}]$, and define $z_{n_{\text{sim}}}^*$ as the lower $\gamma$ quantile of $\min_{j}\widehat\Sigma_{jj}^{-1/2}(\mathbf{c}) \widehat{V}_j^\ell$ across $\ell \in [n_{\text{sim}}]$. \label{alg_line:simulated_z_score}

    \State Calculate the estimated vector of lower confidence bounds:
    $$\widehat{\mathbf{L}}^{n_\text{sim}}(c, c_0) = \hat\tau(\mathbf{c}) - \frac{z_{n_\text{sim}}^*}{\sqrt{|{{D_{\text{test}}}|}}} \hat{\sigma}(\mathbf{c}),  $$
    where $\hat{\sigma}(\mathbf{c})$ is the diagonal entries of $\widehat{\Sigma}^{-1/2}(\mathbf{c})$. 

    \State \textbf{return} $\mathbf{1}[\widehat{\mathbf{L}}^{n_\text{sim}}(\mathbf{c},c_0) > 0]$. 
    \end{algorithmic}
    
    \caption{Safety Test on $D_{\text{test}}$ for Multiple Cutoffs.}
    \label{alg:multiple_safety_test}
\end{algorithm}

The key modification in Algorithm \ref{alg:multiple_safety_test} that enables multiple testing is the adjusted critical value $z^*_{n_{\text{sim}}}$, which takes into account the covariance of the limiting distribution $\hat{\tau}(\mathbf{c}) \sim N(\tau(\mathbf{c}), \Sigma(\mathbf{c}))$. 
With $n_{\text{sim}} = \infty$, $z^*_{\infty}$ corresponds to the minimum adjusted critical value that (i) contains  $1-\gamma$ mass of $\hat{\tau}(c)$'s limiting distribution as $|D_{\text{test}}| \rightarrow \infty$, and (ii) maintains proportional lower confidence bound widths to the standard deviation of policy difference estimates. A direct consequence of (ii) is that Algorithm \ref{alg:multiple_safety_test} requires minimal implementation overhead. The only additional step necessary is the simulation of multivariate normal vectors, which is readily implemented in many statistical software packages, and can be precomputed over the space of candidate policies $\Ccal$. The simulation step in Line \ref{alg_line:simulated_z_score} attempts to approximate $z^*_{\infty}$ with a large value of $n_{\text{sim}}$. We provide the formal result in Proposition \ref{thm:multiple_testing_correctness} below:

\begin{prop}[Asymptotic Correctness of Sup-$t$ band]\label{thm:multiple_testing_correctness}
    Under Assumptions \ref{assump:rubin_po}, \ref{assump:unconfoundedness}, \ref{assump: positivty}, and the conditions of Propostion \ref{thm:asymp_normal_efficient}, as $n_{\text{test}}, n_{\text{sim}} \rightarrow \infty$, we have asymptotically correct joint coverage, i.e.,
    $$\lim_{{n_\text{sim}} \rightarrow \infty}\lim_{n_{\text{test}} \rightarrow \infty} \sup_{P \in \Pcal}\PP_P\left(\tau(c_i) < \widehat{\mathbf{L}}^{n_\text{sim}}_i(c, c_0)\right) \leq \gamma, \quad \forall \ i \in [l],$$
    where $\widehat{\mathbf{L}}^{n_{\text{sim}}}_i(c, c_0)$ is the $i$-th entry of $\widehat{\mathbf{L}}^{n_{\text{sim}}}$ as defined in Algorithm \ref{alg:multiple_safety_test}.  
\end{prop}

Proposition \ref{thm:multiple_testing_correctness} implies that when testing for multiple cutoffs, we still discover a cutoff policy that improves upon the baseline cutoff policy $\pi(c_0)$ if only a single $c_i$ passes the lower bound test.

% \section{Heuristic Approaches for Policy Selection} 
\section{Heuristics  for Cutoff Selection} 
\label{sec:cutoff_selection}

The methods in Section \ref{sec:testing_procedures} give us more powerful, flexible tools to test a given policy cutoff $\pi(c)$. In this section, we focus on 
providing two heuristic algorithms for selecting
cutoffs on $D_{\text{tune}}$, which will be used as inputs for Algorithms \ref{alg:safety_test} and \ref{alg:multiple_safety_test}. 
In Algorithm~\ref{alg:single_selection}, we introduce our single cutoff selection heuristic inspired by the cutoff selection process of \cite{pmlr-v37-thomas15}. 
In Algorithm~\ref{alg:multi_selection},
we introduce our multi-cutoff heuristic, which generalizes the single-cutoff approach by (1) using the estimated probabilities of passing, and (2) adding multiple cutoffs for the safety test consideration.
We present the overall testing procedures in Algorithm~\ref{alg:MDR-HCPI},  naming the resulting methods \textit{Calibrated Safe Policy Improvement} (CSPI) for testing a single cutoff and \emph{CSPI with Multiple Testing} (CSPI-MT) for testing multiple cutoffs.

\subsection{Single Cutoff Selection}
We introduce our heuristic approach for selecting a single cutoff in Algorithm~\ref{alg:single_selection}. 
Our selection criteria
aims to account for the probability of passing, as well as the expected improvement if the new cutoff is adopted.
In particular, we first estimate the probability $\widehat P$ of passing the asymptotic test on $D_{\text{test}}$ while only observing the tuning dataset $D_{\text{tune}}$. 
$\widehat P$ is calculated by using plug-in estimates of the standard deviation and mean of the asymptotic distribution of our estimator $\hat{\tau}$, treating it \emph{as if we had the same number of samples as $D_{\text{test}}$}. In our selection criteria, we multiply the probability of passing with the estimated policy difference, selecting the policy that yields the highest expected improvement.

We note that existing works \cite{pmlr-v37-thomas15} have used similar intuitions for selecting policy cutoffs. For example, Thomas et. al. \cite{pmlr-v37-thomas15} account for the probability of passing the test by rescaling lower confidence bounds and checking whether this value is above 0. In contrast, our approach allows for a \emph{continuous} probability, offering a finer estimate of whether a test is likely to pass. We compare both approaches in Section \ref{sec:empirics}, showing that opting for our power calculation
improves upon both the rate of passing our safety test and the realized improvement. 

\begin{algorithm}[t]
    \begin{algorithmic}[1]
    \State \textbf{Input}: $D_{\text{tune}}$, number of folds $K$,
    model classes for estimating $\hat{\mu}, \hat{e}$, error tolerance $\gamma$, the set of all candidate cutoffs $\Ccal$,
    sample size of test set $|D_{\text{test}}|$.

    \State Estimate vector of estimates and covariance matrix $\hat{\tau}(\Ccal)$, $\widehat{\Sigma}(\Ccal)$ as in Definition \ref{defn:policy_diff_est} using $K$ folds on $D_{\text{tune}}$.

    \State Calculate the estimated probability of passing:
    \begin{equation}\label{eq:prob_passing}
    \widehat{P}(\Ccal, c_0) = [\widehat{P}(c_1, c_0), ..., \widehat{P}(c_G, c_0)], 
    \end{equation}
    where $\widehat{P}(c_i, c_0) = 1-\Phi_{\mu = \hat{\tau}(c_i), \sigma^2 = \frac{\widehat{\Sigma}_{ii}(\Ccal)}{|{D_{\text{test}}}|}}
\left(\sqrt{\frac{\widehat{\Sigma}_{ii}(\Ccal)}{|{D_{\text{test}}|}}}\Phi^{-1}(1-\gamma)\right)$.

    \State \textbf{return} $c^*_1 = \argmax_{c_i \in 
    \Ccal} \widehat{P}(c_i, c_0) \times \hat\tau(c_i)$. 
    \end{algorithmic}
    \caption{Single Cutoff Selection
    on $D_{\text{test}}$.}
    % for a Single Cutoff $c$.}
    
    % \caption{Selection Algorithm on $D_{\text{test}}$ for a Single Cutoff $c$.}
    \label{alg:single_selection}
\end{algorithm}

\subsection{Adaptive Selection of Multiple Policy Cutoffs}\label{subsec:adaptive_multi_cutoffs}

%%% Figure 1
\begin{figure*}[t]
    \centering
    \includegraphics[width=0.33\linewidth]{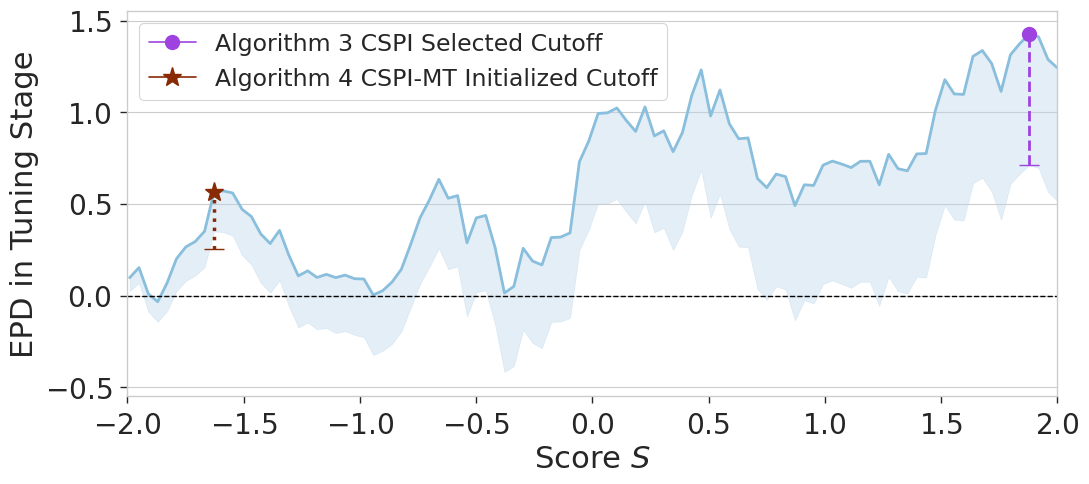}
    \includegraphics[width=0.33\linewidth]{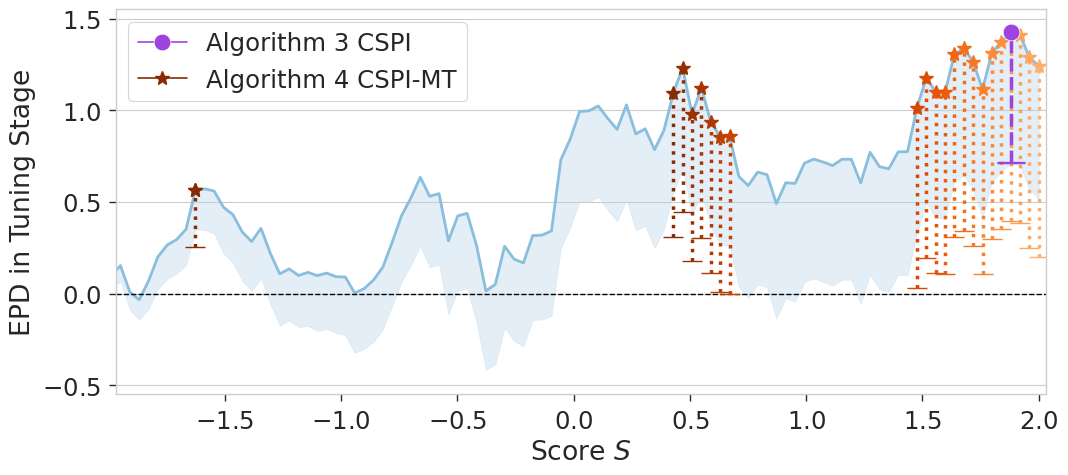}
    \includegraphics[width=0.33\linewidth]{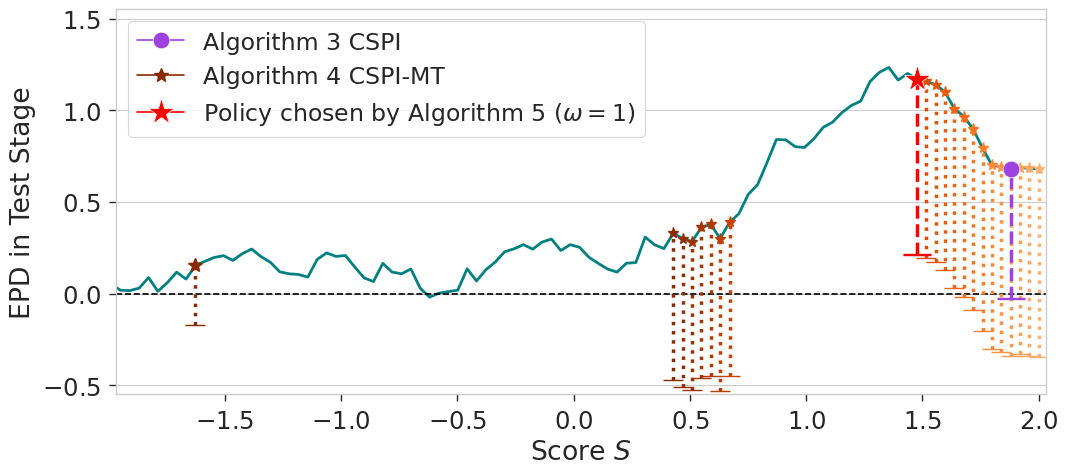}
    \caption{Visual comparison of Algorithms \ref{alg:single_selection} and \ref{alg:multi_selection} on DGP2, defined in Section \ref{sec:empirics}.
    %%%%
    %%%%%%%%%%%%%%%%%%%
    \textit{Left} and \textit{Middle}: selecting cutoff policies on $D_{tune}$.
    % using Algorithms~\ref{alg:single_selection} and~\ref{alg:multi_selection}. 
    EPD: estimated policy value difference(s).
    % During the tuning stage on $D_{tune}$, Algorithm 4 is initialized with the cutoff that maximizes the passing probability, and Algorithm 3 with the maximum expected improvement cutoff. 
    % \textit{Middle}: Algorithm 4 identifies a sequence of cutoffs that improve the expected policy difference (EPD). 
    \textit{Right}: Testing selected policies on $D_{test}$.
    % While the cutoff chosen by Algorithm 3 fails the safety test on $D_{test}$, several of the cutoffs identified by Algorithm 4 both pass the test and improve the policy gain. Figures utilize DGP2 (defined in Sec.~\ref{sec:empirics}) and
    The estimated lower confidence bound for each cutoff is indicated by the corresponding horizontal bar.}
    \label{fig:viz_alg_5}
\end{figure*}

While Algorithm \ref{alg:single_selection} provides a natural, intuitive approach for selecting a \emph{single cutoff}, its performance heavily depends on a simple estimation conducted on only a small fraction of the data.
In practice, choosing a single cutoff often fails to protect against the inherent uncertainty due to the limited size of $D_{\text{tune}}$, where
passing probabilities (or checking whether rescaled lower bounds are above 0) are overly optimistic.
This results in lower rates of the selected policy passing our safety test in noisy, low-signal settings. 

To address this issue, we leverage \emph{multiple testing}  to
maximize the detection of cutoffs that improve significantly over the baseline, without sacrificing detection power.
Our heuristic
begins by selecting
the cutoff policy that is most likely to \emph{pass} (as opposed to the policy with the highest expected improvement in Algorithm~\ref{alg:single_selection}). It then greedily adds cutoffs with larger estimated policy improvement until the lower confidence bound of a previously added point falls below zero. We provide the pseudocode for our multiple testing heuristic in Algorithm \ref{alg:multi_selection}.

\begin{algorithm}[t]
    \begin{algorithmic}[1]
    \State \textbf{Input}: $D_{\text{tune}}$, number of folds $K$, model classes for $\hat{\mu}, \hat{e}$, error tolerance $\gamma$, the set of all candidate cutoffs $\Ccal$,
    sample size of test set $|D_{\text{test}}|$, number of simulations for critical value $n_{\text{sim}}$. 

    \State Estimate vector of estimates and covariance matrix $\hat{\tau}(\Ccal)$, $\widehat{\Sigma}(\Ccal)$ as in Definition \ref{defn:policy_diff_est} using $K$ folds on $D_{\text{tune}}$.

    \State Calculate the estimated probability of passing $\widehat P(\Ccal, c_0)$ according to  Equation~\eqref{eq:prob_passing}.

    \State Calculate $c_1^* = \argmax_{c \in \Ccal} \widehat{P}(c, c_0)$.
    \State Initialize 
    the set of policies $\mathbf{c}_{\text{test}}=\{c^*_1\}$.

    \State Calculate the set of eligible candidates,
    $\mathbf{c}_{\text{cand}} = \{c \in \Ccal: \hat\tau(c) \geq \hat{\tau}(c^*_1)\}$,
    and simulate i.i.d. normal vectors $\widehat{V}^\ell \sim N(\mathbf{0}, \widehat{\Sigma}(\mathbf{c}_{\text{cand}}))$.

    \State Calculate the correlation vector for 
    the policy difference of
    each candidate cutoff relative
    to the initial cutoff $c_1^*$:
    $$\hat{\eta}(\mathbf{c}_{\text{cand}},  c_1^*) = \{\hat\eta(c, c_1^*)\}_{c \in \mathbf{c}_{\text{cand}}}, $$
    where $\hat\eta(c_i, c_1^*) = |\frac{\widehat\Sigma_{c_i, c_1^*}(\Ccal)}{\sqrt{\widehat\Sigma_{c_i, c_i}(\Ccal)\widehat\Sigma_{c_1^*, c_1^*}(\Ccal)}}|$.

    \State Sort elements in $\mathbf{c}_{\text{cand}}$ in descending order of $\hat{\eta}(\mathbf{c}_{\text{cand}}, c_1^*)$.
    \For{$c \in \mathbf{c}_{\text{cand}}$}
    \State $\mathbf{c}_{\text{temp}} = \mathbf{c}_{\text{test}} \cup c$.
     \State Calculate the adjusted critical value $z^{\text{temp}}_{n_{\text{sim}}}$ as the lower $\gamma$ quantile of $\min_{\mathbf{c}_{\text{temp}}} \widehat\Sigma_{c,c}^{-1/2}(\Ccal) \widehat{V}_j^\ell$ across $\ell \in [ n_{\text{sim}}]$. 
     \State Construct lower bound vector $\widehat{\mathbf{L}}^{n_{\text{sim}}}_{\text{temp}}(\mathbf{c}_{\text{temp}}, c_0) $:
     $$\widehat{\mathbf{L}}^{n_{\text{sim}}}_{\text{temp}}(\mathbf{c}_{\text{temp}}, c_0) = \hat\tau(\mathbf{c}_{\text{temp}}) - \frac{z_{n_\text{sim}}^\text{temp}}{\sqrt{|{D_{\text{test}}}|}} \hat{\sigma}(\mathbf{c}_{\text{temp}}). $$

     \State If all entries of $\widehat{\mathbf{L}}^{n_{\text{sim}}}_{\text{temp}}(\mathbf{c}_{\text{temp}}, c_0)$ lie above 0, set $\mathbf{c}_{\text{test}} =\mathbf{c}_{\text{temp}}$.

    \EndFor

    \State \textbf{return} 
    % the selected set of 
    cutoff policies set
    $\mathbf{c}_{\text{test}}$. 
    \end{algorithmic}
    \caption{Multiple Cutoff Selection 
    on $D_{\text{test}}$.}
    
    \label{alg:multi_selection}
\end{algorithm}

\begin{algorithm}[t]
    \begin{algorithmic}[1]
    \State \textbf{Input}: Data $\{O_i\}_{i=1}^n$, error tolerance $\gamma$, split ratio $\zeta$, number of folds $K$, model classes for $\hat{\mu}$, $\hat{e}$, candidate cutoffs $\Ccal$,
    and 
    $n_{\text{sim}}$, multiple test parameter $\omega \in \{0,1\}$.

    \State Randomly split $\{O_i\}_{i=1}^n$, allocating $\zeta$ fraction of the samples to $D_{\text{tune}}$, and the remainder to $D_{\text{test}}$.

    \State \textbf{Step 1}: 
    Use Algorithm~\ref{alg:single_selection} if $\omega = 0$ (else use Algorithm \ref{alg:multi_selection}) with $D_{\text{tune}}$  
    to obtain the corresponding cutoff set,
    $\mathbf{c}_{\text{test}}$. 

    \State \textbf{Step 2}: Use Algorithm~\ref{alg:safety_test} if $\omega=0$ (else use Algorithm \ref{alg:multiple_safety_test}) with $D_{\text{test}}$
    to decide which cutoff(s) pass the safety test, and denote the resulting cutoff set $\mathbf{c}_{\text{pass}}$. 

    \If {$|\mathbf{c}_{\text{pass}}| = 0$}
    \textbf{return} $c_0$.
    \ElsIf{$\omega$ = 0} \textbf{return} $\mathbf{c}_{\text{pass}}$.
    \ElsIf{$\omega = 1$}, estimate $\hat{\tau}(c)$ for all $c \in \mathbf{c}_{\text{pass}}$
    using $K$ folds on the entire dataset $\{O_i\}_{i=1}^n$, and  \textbf{return} $c^*_1 = \argmax_{c \in \mathbf{c}_{\text{pass}}} \hat{\tau}(c) $.
    \EndIf

    \end{algorithmic}
    
    \caption{CSPI and CSPI-MT.}
    \label{alg:MDR-HCPI}
\end{algorithm}

\begin{remark}[Computational Justification for Our Heuristic]
While 
an exhaustive search over the 
candidate cutoff set can identify those
most likely to pass, 
such an approach scales exponentially
with the size of 
our candidate set because
the critical value, $z_{n_{\text{sim}}}^\text{temp}$, depends on both the number 
and the values of the
cutoff being tested. 
Our greedy heuristic, however,
scales linearly with
the size of
our candidate set, 
reducing computation time for selecting test candidates.

\end{remark}

%%% Figure 2

\begin{figure*}[htp]
  \centering
  \includegraphics[scale=0.15]{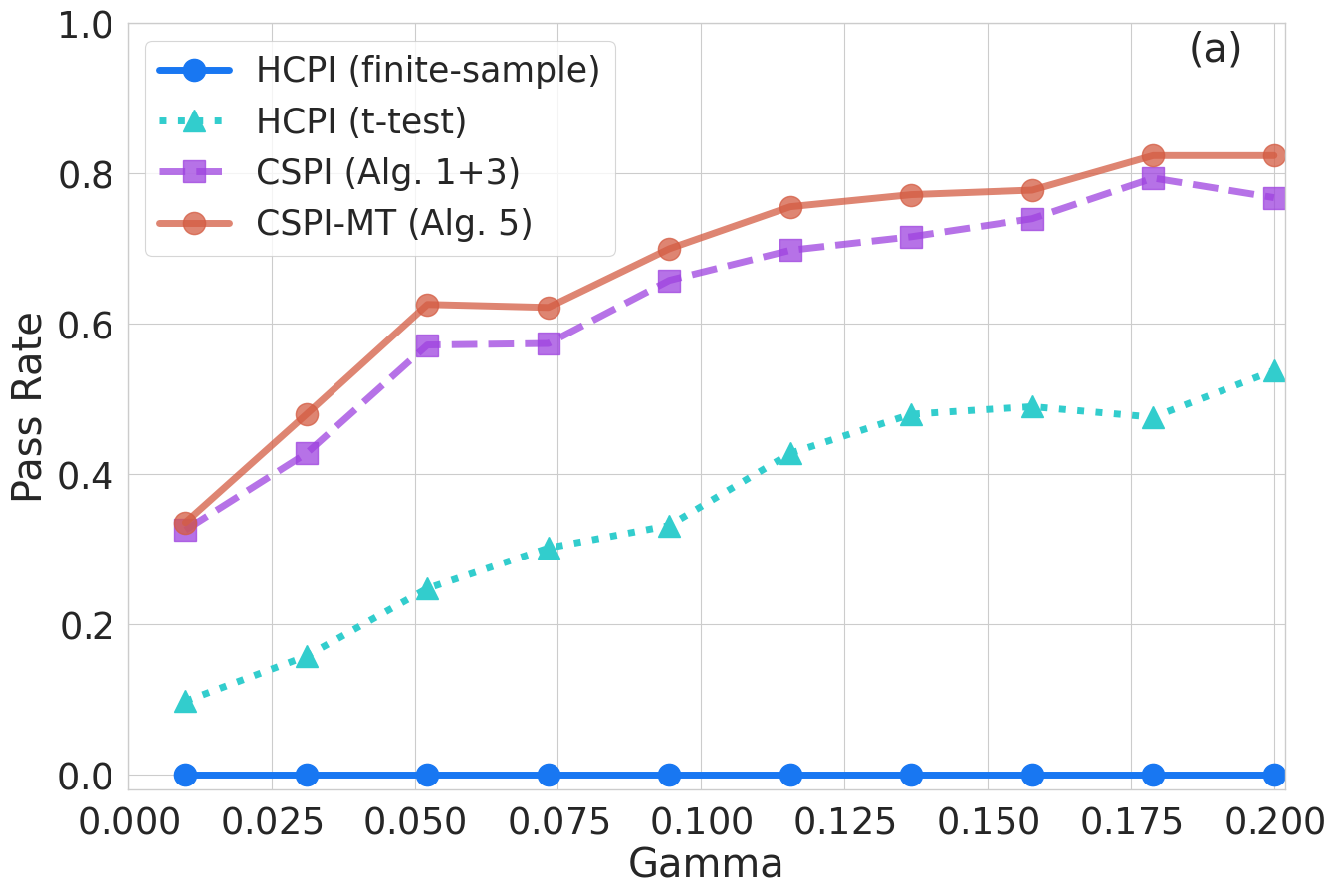}
  \includegraphics[scale=0.15]{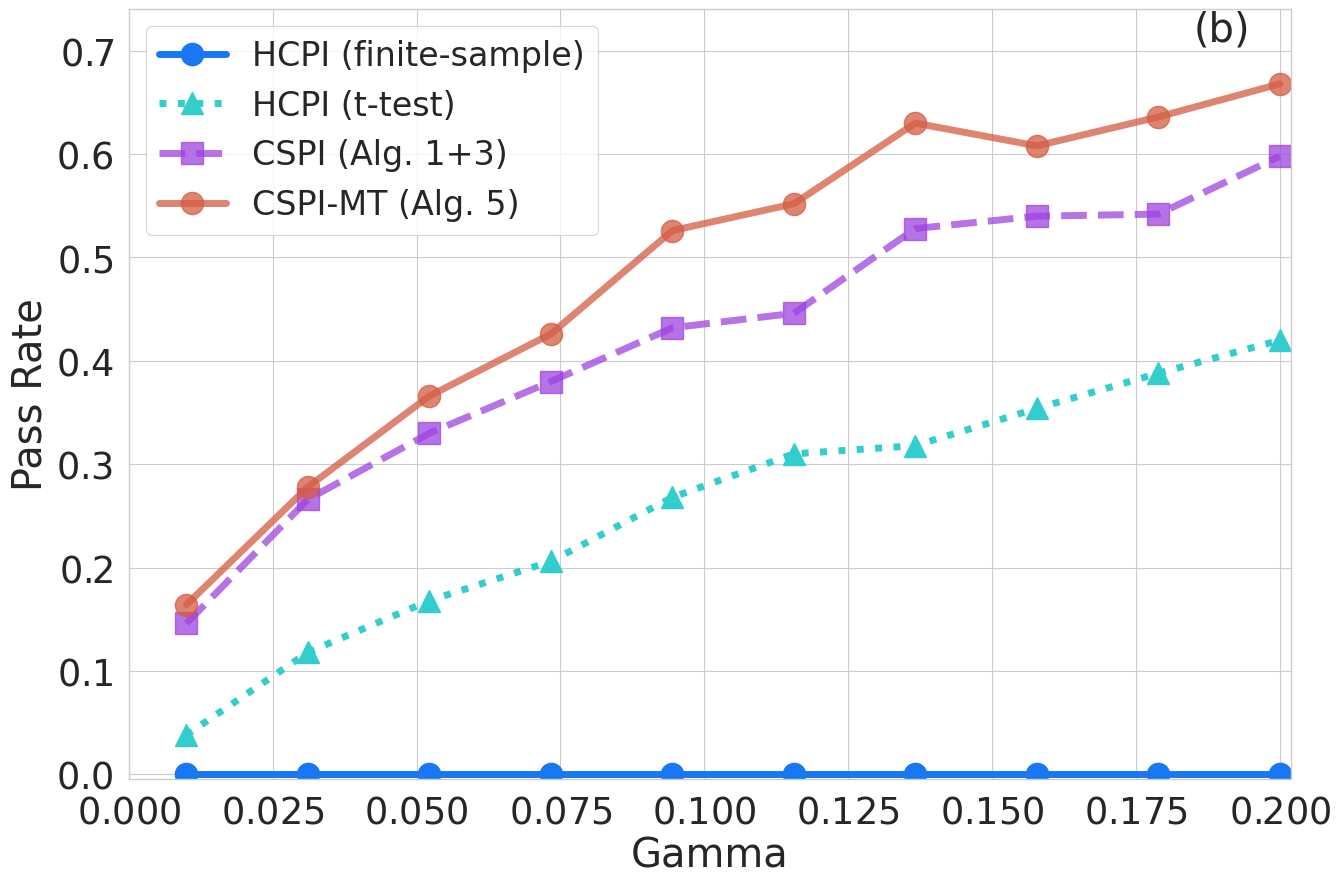}
  \includegraphics[scale=0.15]{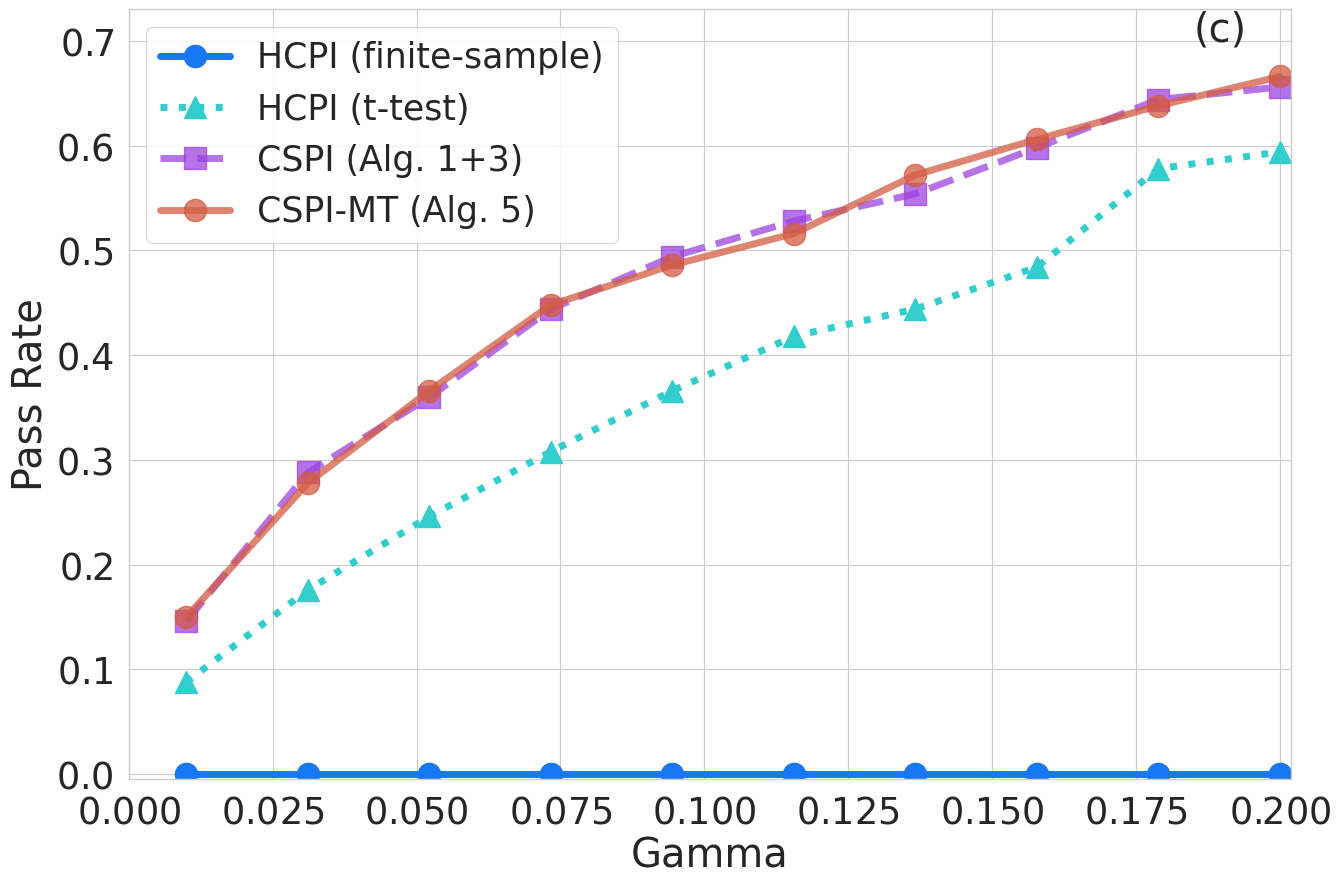}
  \hfill
  \includegraphics[scale=0.15]{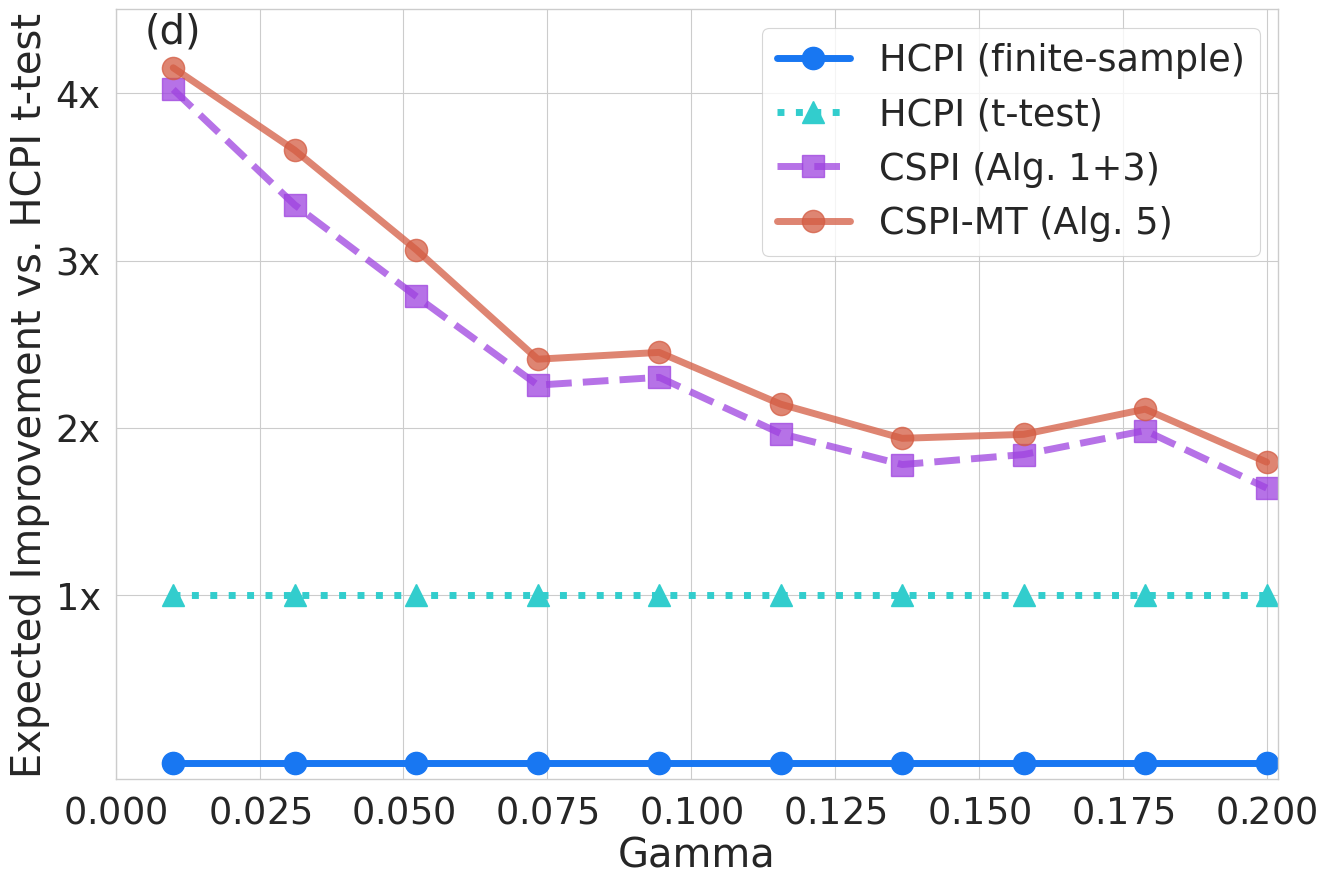}
  \includegraphics[scale=0.15]{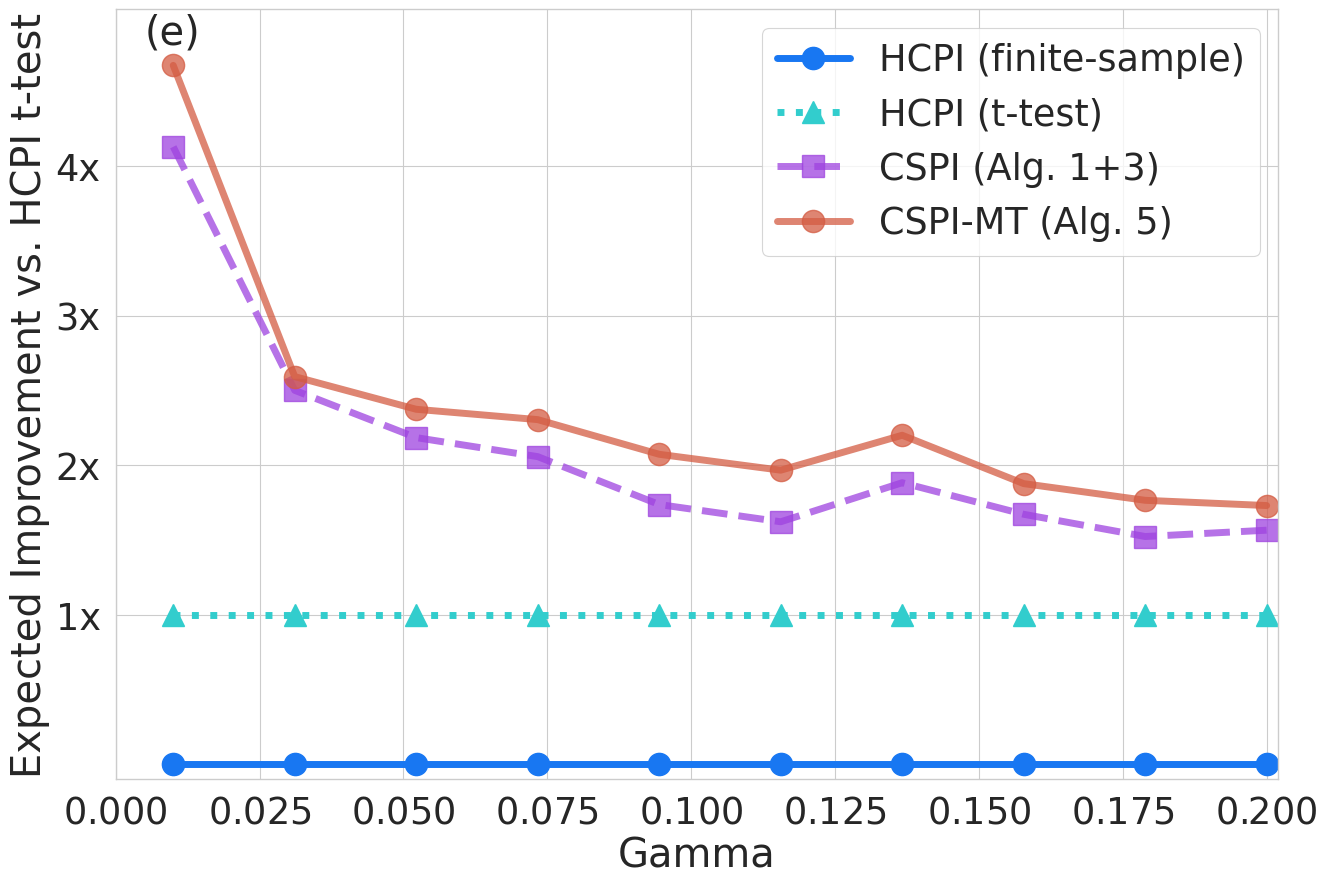}
  \includegraphics[scale=0.15]{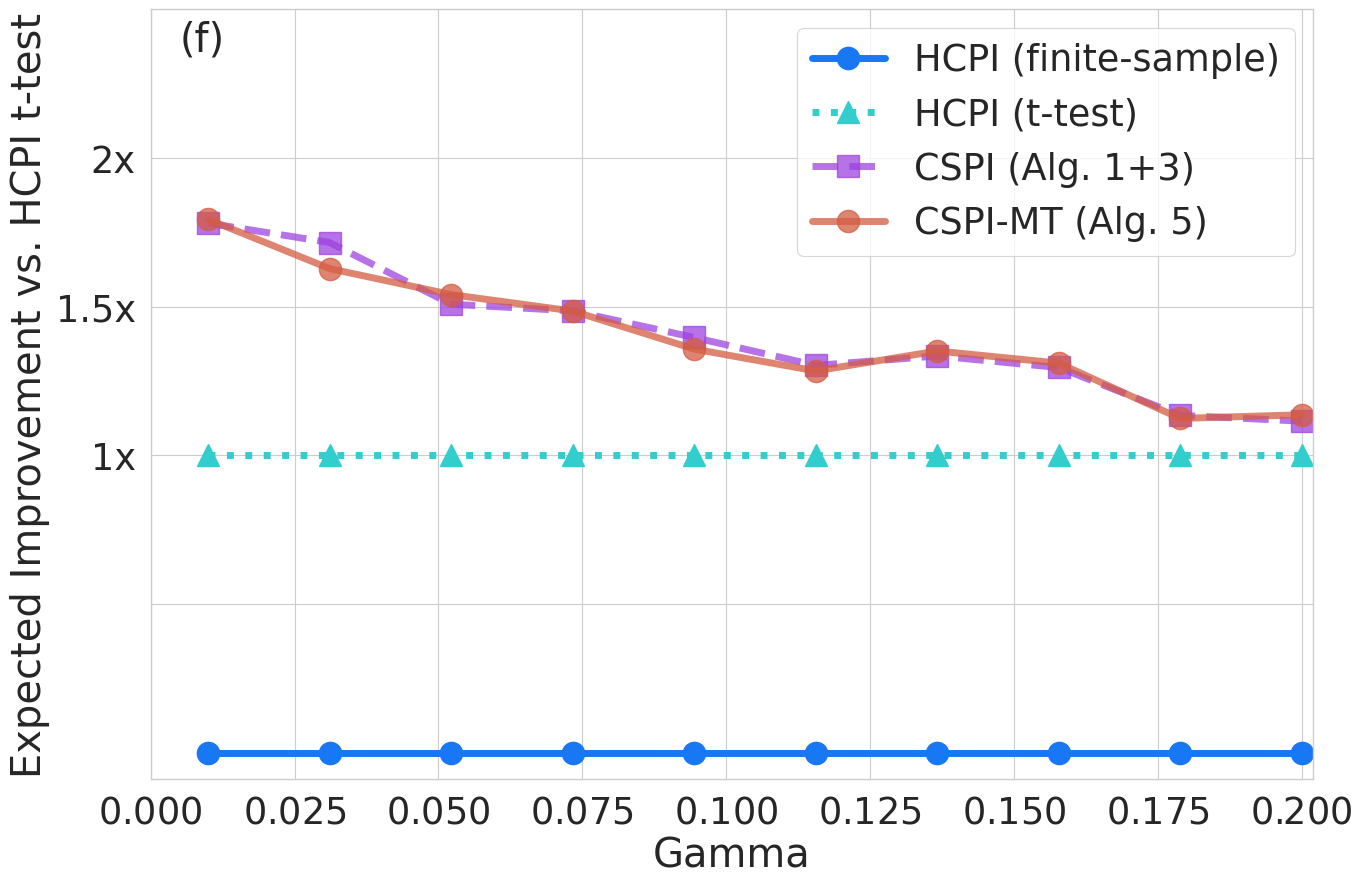}
\caption{Pass rate of selected cutoffs and expected improvement across $\gamma$ values. \textit{Top row}: Pass rates from left to right for DGP1, DGP2, JOBS (with baseline: treat-none). \textit{Bottom row}: Expected improvement compared to HCPI (t-test) for DGP1, DGP2, JOBS.}
\label{fig:pass_ei_plots}
\end{figure*}

We introduce 
our 
calibrated safe policy improvement with multiple testing
algorithm in Algorithm \ref{alg:MDR-HCPI}. Algorithm \ref{alg:MDR-HCPI} provides a general framework for selecting either one cutoff (CSPI, constructed using Algorithms \ref{alg:safety_test} and \ref{alg:single_selection}) or adaptively selecting multiple cutoffs (CSPI-MT, constructed using Algorithms \ref{alg:multiple_safety_test} and \ref{alg:multi_selection}). If opting for multi-cutoff testing, the last step of Algorithm \ref{alg:MDR-HCPI} picks among the cutoffs that passed the safety test using the entire dataset.
This is intended to obtain the \emph{most accurate} estimate of the policy gains. This does not affect the safety guarantee provided by Proposition \ref{thm:multiple_testing_correctness}, as our safety tests are \emph{only} selected on the data $D_{\text{tune}}$.

Figure \ref{fig:viz_alg_5} 
contrasts
Algorithm~\ref{alg:MDR-HCPI} under single and multiple testing settings.
Figure \ref{fig:viz_alg_5} (left) compares the single cutoff (that maximizes the estimated expected improvement) selected by Algorithm~\ref{alg:single_selection} with the initial cutoff (that maximizes the expected probability of passing) selected by Algorithm~\ref{alg:multi_selection} on $D_{\text{tune}}$. The curves represent the estimated policy differences across cutoff values with the baseline of $c_0=-2$. The shaded areas indicate the estimated lower confidence bounds calculated according to Algorithm~\ref{alg:single_selection}.
Figure \ref{fig:viz_alg_5} (middle) further illustrates the cutoff selection process described in Lines 8-12 of
Algorithm~\ref{alg:multi_selection}, with darker colors indicating cutoffs selected closer to the initial cutoff. The added cutoffs have higher estimated policy gains than the initial cutoff $c_0$ (Line 6), and all estimated lower bounds are above 0 (Line 11). Note that the estimated lower bounds for Algorithm~\ref{alg:multi_selection}'s selected cutoffs (shown by the horizontal bars) are lower than the shaded region, showing the 
% necessary 
correction made by Algorithm~\ref{alg:multi_selection} to ensure statistically valid multiple testing. 

Figure \ref{fig:viz_alg_5} (right) shows the results of conducting safety tests using Algorithm~\ref{alg:MDR-HCPI} on $D_{\text{test}}$ (calculated by using Algorithms \ref{alg:safety_test} and \ref{alg:multiple_safety_test} for the cutoffs selected by Algorithms \ref{alg:single_selection} and \ref{alg:multi_selection}, respectively). The teal curve represents the estimated policy value. The cutoff chosen by Algorithm \ref{alg:single_selection} fails to pass the safety test, as its $1-\gamma$ lower confidence bound falls below 0. In contrast, Algorithm \ref{alg:multi_selection} obtains at least one cutoff that passes the safety test of Algorithm \ref{alg:multiple_safety_test}, allowing a policy change with probability $1-\gamma$ of improving over the baseline.

\section{Empirical Results} \label{sec:empirics}
In this section, we empirically validate CSPI and CSPI-MT against existing policy improvement approaches. We show the benefits of our approaches for (1) improved detection rates for cutoff policies with higher policy value than the baseline, (2) larger average policy improvements, and (3) calibrated error control in adversarial settings where no policies are better than the baseline. 
We begin with our synthetic datasets, in which we vary the treatment effect size with respect to the score. 
We then discuss a semi-synthetic example using the 
French job-search counseling service dataset \cite{behaghel}. 

For both our synthetic and semi-synthetic results, we use $n_{\text{sim}} = 10,000$ for the calculation of our adjusted $z$-score, utilize 20\% of the data for $D_{\text{tune}}$, and 80\% of the data of $D_{\text{test}}$.\footnote{This is the same split fraction $\zeta$ used in HCPI~\cite{pmlr-v37-thomas15}.} 
In all datasets, the propensity scores are constant across users but unknown, reflecting data collected from randomized experiments without recording the propensity scores. We set the propensity
score $\hat{e}$ using the sample mean of observed treatment, i.e., $\frac{1}{n}\sum_{i=1}^n A_i$.

\paragraph{Baseline Methods of Comparison}

We compare our two heuristics in Algorithm \ref{alg:MDR-HCPI},  CSPI (for testing a single cutoff) and CSPI-MT (for testing multiple cutoffs), with the following procedures that satisfy the asymptotic safety property in Definition \ref{defn:asymp_safety}:
\begin{itemize}[left=1pt]
    \item HCPI (finite-sample): HCPI procedure with concentration-inequality based test (modified Algorithm 1 in \cite{pmlr-v37-thomas15}). 
    \item HCPI ($t$-test based): HCPI procedure with importance-weighted asymptotic test (modified Algorithm 2 in \cite{pmlr-v37-thomas15}).
\end{itemize}
In particular, we make minor modifications to the existing HCPI algorithms to ensure that our error guarantees hold for policy differences, rather than policy values. We provide  details on these modified methods in Appendix \ref{sec:appendix_policy_diff}.

\subsection{Experiment Setup}

\paragraph{Synthetic Simulation Settings} 
We use three DGPs  for our simulation examples. For each DGP, $X = (S, X_1, X_2) \in \RR^3$, with $S \in \text{Unif}([-2,2])$ and $X_1, X_2 \sim N(0,25)$. For DGP $i$, the true conditional outcomes functions $\mu_{i,0}, \mu_{i,1}$ are given by:
$$\mu_{i,0} = 2X_1 - 3X_2 + \epsilon, \quad \mu_{i,1} = f_i(S) - 2X_1 - 3X_2 + \epsilon, $$
$$f_1(S) = S, \ f_2(S) = 4*\mathbf{1}[S \geq 1.5] - 2* \mathbf{1}[S \leq -1.5], \ f_3(S) = -0.25,$$
with $\epsilon\sim N(0,25)$. The three DGPs correspond to different relationships between the score $S$ and the underlying treatment effect distribution: (i) DGP1: $S$ is linearly related with user treatment effects, (ii) DGP2: $S$ correctly orders user treatment effects, (iii) DGP3: $S$ is uncorrelated with user treatment effects. For all DGPs, $e(X) = 1/2$ (equal randomization in control and treatment). For DGP1 and DGP3, we use the baseline of treat none ($c=2$) since the average treatment effect is 
less than or equal to 0. 
For DGP2, we use the baseline of treat all ($c=-2$) as the treatment effect is positive. 
These baselines represent scenarios where decision-makers naively decide whether to implement the treatment across the entire population based solely on the average treatment effect. 
To satisfy the conditions in Proposition \ref{thm:asymp_normal_efficient},
we 
correctly specify the regression model class for
$\hat{\mu}$ 
to be linear regression. 
We test 10 and 20 uniformly spaced $\gamma$ values between $[0.01, 0.2]$ for DGP1/DGP2 and DGP3, respectively. We perform 500 simulations per $\gamma$. Across all simulations, the outcome variance is much larger than the treatment effect, corresponding to a low signal-to-noise regime.

\paragraph{Semi-Synthetic Simulation Settings} The French jobs dataset contains 33,797 observations of unemployed individuals participating in a large-scale randomized experiment comparing assistance programs \cite{behaghel}. We denote $X$ as the demographic variables recorded for each individual and $Y$ as the (binary) outcome of reemployment in six months. The control group ($A=0$) receives a privately-run counseling program, while the treatment group ($A=1$) receives a publicly-run counseling program. 
In our hypothetical scenario, we set the score $S$ to be the age of an individual.
To represent the status-quo, we set the baseline policy as treat-none. To construct our semi-synthetic dataset, we bootstrap sample $25,000$ observations with replacement according to the sample weights provided in the dataset. 
Since the ground truth data generating mechanism is unknown on this dataset, we use (nonparametric) gradient boosted classifiers
to estimate $\hat{\mu}$.
We set the maximum iterations to be 20 to prevent overfitting, with early stopping. The true unknown policy differences are estimated with a bootstrap sample size equivalent to the size of the entire dataset for Figure \ref{fig:pass_ei_plots}. We refer to this bootstrap, semi-synthetic DGP as JOBS.

\subsection{Discussion of Empirical Results}

\subsubsection{Improved Detection Power and Policy Improvement.}
For DGP1, DGP2, and JOBS, both CSPI and CSPI-MT drastically outperform HCPI methods in detecting policies better than the baseline. 
Figures \ref{fig:pass_ei_plots}(a), \ref{fig:pass_ei_plots}(b), and \ref{fig:pass_ei_plots}(c) show the empirical rate of the selected cutoff(s) passing the safety test with different $\gamma$ values for DGP1, DGP2, and JOBS, respectively. 
Due to poor calibration, HCPI (finite-sample) fails to detect any cutoffs better than the baseline, despite the fact that at least half of the cutoff candidates improve over the baseline policy value. 
While HCPI ($t$-test) improves upon the detection rates and the estimated expected improvement over HCPI (finite-sample), 
its overly wide confidence intervals result in lower pass rates and improvements when compared to our approaches. 
With $\gamma=0.05$, CSPI and CSPI-MT achieve more than double the pass rates of \mbox{HCPI ($t$-test)} on DGP1 and DGP2 and show a 50\% increase on JOBS. 

Building on tests using the minimum-width lower bounds,  both of our heuristics (CSPI, CSPI-MT)
have improved detection rates. We note that the improvement of CSPI and CSPI-MT over \mbox{HCPI ($t$-test)} is largely moderated by predictive performance/convergence rate of our estimates $\hat{\mu}$, $\hat{e}$. 
In our synthetic examples, when used with correctly specified models converging at the rate of $O(n^{-1/2})$, we see more dramatic improvements in both detection power and expected improvement relative to the JOBS dataset. 
This is to be expected: correctly specified linear regression models converge at faster rates to the underlying regression model compared to complex, machine-learning approaches such as random forests \cite{chernozhukov2017doubledebiased}. 

\begin{figure}[h]
   \centering
     \includegraphics[width=0.75\linewidth]{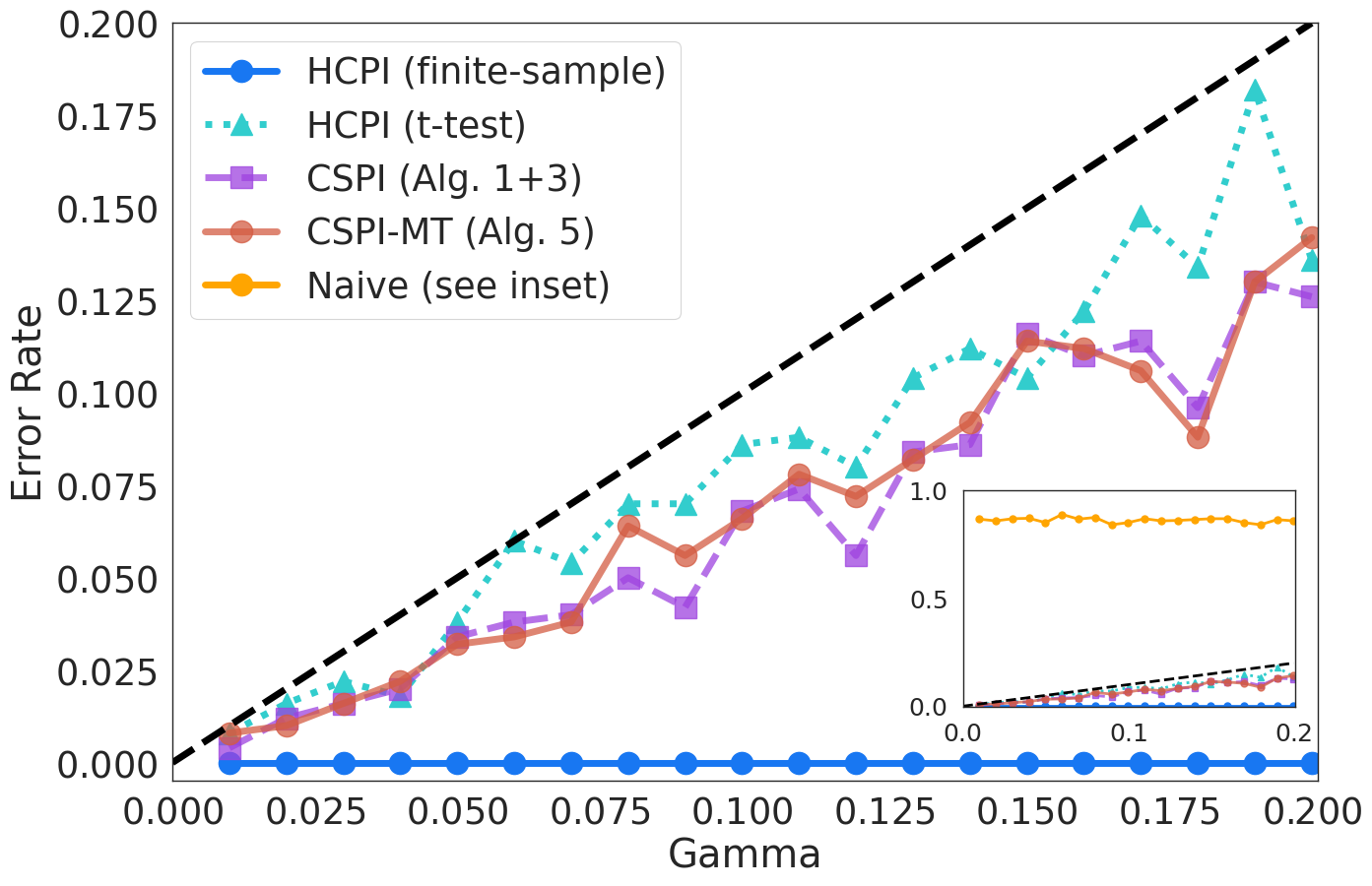}
   \caption{Error Rates for DGP3, with $n=2000$.}
   \Description{description}
   \label{fig:error_rate_DGP3}
\end{figure}
The relative expected improvement plots in Figures~\ref{fig:pass_ei_plots}(d), \ref{fig:pass_ei_plots}(e), and \ref{fig:pass_ei_plots}(f) show that the cutoffs passing the safety test of CSPI and \mbox{CSPI-MT} provide policy improvements over the baseline policy. At the smallest $\gamma$ values tested, both CSPI and CSPI-MT provide at least  $4\times$ the expected improvement compared to HCPI ($t$-test), and uniformly over all other $\gamma$ values. 
CSPI-MT achieves the largest increase in detection and policy improvement under DGP2 due to:
(i) the weak correlation between the score variable and user effects;
(ii) the presence of large regions of users 
(defined with respect to $S$) with null treatment effects;  and 
(iii) relatively smaller potential gains compared to DGP1.

% \vspace{-2mm}
\subsubsection{Calibration of Error Rates in Moderate Sample Sizes}\label{empirics:calibration}
Since the calibration guarantee of our algorithms holds only in asymptotic settings,
our 
procedures may have inflated error rates in the small sample settings where the baseline policy achieves the highest policy value.
To this end, 
we test our approaches on DGP3 with a baseline of treating none (the optimal policy) using only $n=2000$ samples.
Figure \ref{fig:error_rate_DGP3} shows the specified error rate $\gamma$ against the actual error rate of each approach. 
Selecting the naïve policy cutoff corresponding to the largest estimated policy gain 
leads to an error rate $>80\%$. This demonstrates the value of our safety constraints: HCPI, CSPI, and CSPI-MT all achieve error rates below the specified $\gamma$ across the tested $\gamma$ range. Even with moderate sample sizes and an optimal baseline, all asymptotic procedures maintain error rates within the specified $\gamma$-level and achieve 
similar
calibration error
compared to HCPI (finite-sample). 
This shows that the improved detection rates of CSPI and CSPI-MT 
do not come at the cost of violating 
the specified $\gamma$-level error constraint, offering similar error control to
HCPI (t-test).

% \vspace{-4mm}
\section{Conclusion and Discussions} \label{sec:conclusion}
In this paper, we propose CSPI-MT and CSPI, two novel 
heuristics that (1) calibrate error rates to the desired safety guarantee, (2) improve the detection rate of policies better than our baseline, and (3) provide increased policy gains relative to existing methods. We conclude with practical takeaways on the methods presented in this work.
CSPI is a computationally lightweight, well-performing option that improves over existing approaches uniformly across experiments and for all specified error-level guarantees. CSPI-MT, our algorithm with the multiple testing heuristic, further improves upon CSPI under settings where additional robustness in cutoff selection is desired. Such settings are most common in low signal-to-noise regimes when the score variable is loosely correlated with user-level treatment effects and when large proportions of the population do not respond strongly to the treatment. 

\newpage
% \newpage
\bibliographystyle{plain}
\bibliography{citation}

\newpage

\appendix

\section{Proofs} \label{sec:appendix_proofs}

We provide brief proofs for Propositions \ref{thm:asymp_normal_efficient} and \ref{thm:multiple_testing_correctness}. 

For a candidate cutoff $c \in \Ccal$ and baseline cutoff $c_0$, define a treatment $A' = \mathbf{1}[\pi(c) \neq \pi(c_0)]$. Note that under Assumption (iii) of Proposition \ref{thm:asymp_normal_efficient}, this is bounded above 0, and by Assumption \ref{assump: positivty}, we trivially obtain that in the region of the covariate space $\min(c, c_0) \leq S \leq \max(c, c_0)$, we have local positivity. Assumptions (i), (ii) of Proposition \ref{thm:asymp_normal_efficient} ensure that Assumptions 3.1 and 3.2 of \cite{chernozhukov2017doubledebiased} hold, such that we obtain valid joint confidence regions. 

Note that the $\hat{\psi}^c(X_i, A_i, Y_i, \hat{\mu}_k, \hat{e}_k)$ are the same as the influence functions used for the average treatment effect estimator in Theorem 5.1 of \cite{chernozhukov2017doubledebiased}, with the first additional term to match the direction of the cutoff change and an indicator term to denote the region over which our policies $\pi(c), \pi(c_0)$ disagree. By Theorem 5.1 of \cite{chernozhukov2017doubledebiased}, we directly obtain Proposition \ref{thm:asymp_normal_efficient} with our reparameterization. 

Proposition \ref{thm:multiple_testing_correctness} follows from Theorem 3.1, 3.2, of \cite{chernozhukov2017doubledebiased}, which states that our cutoff estimate vector $\hat{\tau}(\mathbf{c})$ (for some discrete set of cutoffs $\hat{c}$ satisfying Assumptions \ref{assump:rubin_po}, \ref{assump:unconfoundedness}, \ref{assump: positivty} and all conditions in Prop. \ref{thm:asymp_normal_efficient}) satisfies the limiting distribution $\sqrt{n}(\hat{\tau}(\mathbf{c}) - \tau(\mathbf{c})) \rightarrow N(0, \Sigma)$. Then, by direct application of Equations 5,6 of \cite{sup_t_band}, we obtain Proposition~\ref{thm:multiple_testing_correctness}.

\section{Adapting HCPI Approaches for Policy Differences} \label{sec:appendix_policy_diff}

For all HCPI methods, we use importance-weighted returns $\tilde\psi(X_i, A_i, Y_i)$ for each observation, which take the following form for cutoff policy $\pi(c)$ and baseline policy $c_0$:
\begin{align}
        &\tilde{\psi}_1^c(X_i, A_i, Y_i, e) = Y_i \frac{A_i}{{e}(X_i)},\\
        &\tilde{\psi}_0^c(X_i, A_i, Y_i, e) = Y_i\frac{1-A_i}{1-{e}(X_i)},\\
        &\tilde\psi^c(X_i, A_i, Y_i, e) = \left(1-2(\mathbf{1}[c_0 \geq c])\right) \times \\  &\quad\qquad\qquad\qquad\qquad \mathbf{1}[S_i \in (\min(c_0, c), \max(c_0, c))] \times \label{eq:SinRange} \\
        &\qquad\qquad\qquad
        \left(\tilde{\psi}^c_0(X_i, A_i, Y_i, e) - \tilde{\psi}^c_1(X_i, A_i, Y_i, \hat{\mu}_k, \hat{e}_k)\right).\label{eq:influence_fnc}
    \end{align}
Note that the empirical average of $\tilde\psi^c(X_i, A_i, Y_i)$ across $(X_i,A_i,Y_i)_{i=1}^n$ is an unbiased estimate of $\tau(c)$, i.e. $\EE[\tilde\psi^c(X_i, A_i, Y_i)] = \tau(c)$, as long as $e$ is the true propensity score.

For finite sample HCPI, we use the standard one-sided Hoeffding concentration inequality:
\begin{align*}
    \PP_P\big(\frac{1}{n}\sum_{i=1}^n \tilde\psi(X_i, A_i, Y_i) - \sigma_{{\tilde\psi}}\sqrt{\frac{2\log(1/\gamma)}{n}}\geq  \tau(X_i, A_i, Y_i)\big) \leq \gamma.
\end{align*}

For DGP1, DGP2, DGP3, $\sigma_{\tilde\psi} \leq 66$, which we use to construct finite-sample lower bounds. For JOBS, we use the estimated (constant) propensity score over the entire dataset to obtain $\sigma_{\tilde\psi} \leq 5.6$, which we use to conduct finite-sample HCPI. 

The asymptotic $t$-test HCPI using policy differences uses $\tilde\psi(X_i, A_i, Y_i, e)$ in place of $\hat{\psi}(X_i, A_i, Y_i, \hat{\mu}_k, \hat{e}_K)$ in Equations 6 and 7 of Section \ref{defn:policy_diff_est} for a single cutoff $c$ to construct its estimated policy gain and variance estimates with estimated propensity scores. The cutoff selection and safety test conducted remain unchanged from \cite{pmlr-v37-thomas15}.

\end{document}